\documentclass[conference]{IEEEtran}
\IEEEoverridecommandlockouts
\usepackage{cite}
\usepackage{amsmath,amssymb,amsfonts}
\usepackage{algorithmic}
\usepackage{graphicx}
\usepackage{textcomp}
\usepackage{xcolor}
\usepackage{comment}
\usepackage{soul}
\usepackage{multirow}
\usepackage{array}
\usepackage{hhline}
\usepackage{titlesec}
\usepackage{url}
\usepackage{verbatim}
\usepackage{tikz}
\mathchardef\mhyphen="2D
\usepackage{pgfplots, pgfplotstable}
\usepackage{authblk}

\usepackage{color}
\usepackage{microtype}
\usepackage{wrapfig}
\usepackage{booktabs} 
\usepackage{dsfont}
\usepackage{pgfplots}
\usetikzlibrary{patterns}
\usepgfplotslibrary{fillbetween}
\usepackage{pgfkeys}
\usepackage{lipsum}
\usepackage{mwe}
\usepackage{tikz}
\usetikzlibrary{backgrounds}
\usepackage[style=base]{caption}
\usetikzlibrary{intersections}
\usetikzlibrary{shapes.geometric}
\usetikzlibrary{spy}

\usepackage{pgfplotstable}
\pgfmathdeclarefunction{gauss}{2}{%
  \pgfmathparse{1/(#2*sqrt(2*pi))*exp(-((x-#1)^2)/(2*#2^2))}%
}
\pgfplotsset{compat=1.11,
    /pgfplots/ybar legend/.style={
    /pgfplots/legend image code/.code={%
       \draw[##1,/tikz/.cd,yshift=-0.25em]
        (0cm,0cm) rectangle (3pt,0.8em);},
   },
}
\usepackage{colortbl}
\usepgfplotslibrary{colormaps}
    \def\addlegendimage{\csname pgfplots@addlegendimage\endcsname}

\pgfplotsset{ 
cycle list={%
{draw=black,mark=star,solid},
{draw=black, mark=square,solid}}}
\usepackage{breqn}

\usepackage[a4paper, total={185mm,239mm}]{geometry}

\def\BibTeX{{\rm B\kern-.05em{\sc i\kern-.025em b}\kern-.08em
    T\kern-.1667em\lower.7ex\hbox{E}\kern-.125emX}}

\newcommand\blfootnote[1]{%
  \begingroup
  \renewcommand\thefootnote{}\footnote{#1}%
  \addtocounter{footnote}{-1}%
  \endgroup
}

\usepackage{caption}
\usepackage{fancyhdr}
\fancypagestyle{firstpage}
{
    \fancyhead[L]{\footnotesize © 2023 IEEE.  Personal use of this material is permitted.  Permission from IEEE must be obtained for all other uses, in any current or future media, including reprinting/republishing this material for advertising or promotional purposes, creating new collective works, for resale or redistribution to servers or lists, or reuse of any copyrighted component of this work in other works. This paper is accepted at the 28th IEEE European Test Symposium (ETS) 2023.}
    \fancyhead[R]{}
}

\begin{document}

\title{DeepVigor: \underline{V}ulnerab\underline{I}lity Value Ran\underline{G}es and Fact\underline{OR}s for \underline{D}NNs' Reliability Assessment}

\author[1]{Mohammad Hasan Ahmadilivani}
\author[1]{Mahdi Taheri}
\author[1]{Jaan Raik}
\author[1,2]{Masoud Daneshtalab}
\author[1]{Maksim Jenihhin}
\affil[1]{Tallinn University of Technology, Tallinn, Estonia}
\affil[2]{Mälardalen University, Västerås, Sweden}
\affil[1]{\{mohammad.ahmadilivani, mahdi.taheri, jaan.raik, maksim.jenihhin\}@taltech.ee}
\affil[2]{masoud.daneshtalab@mdu.se}

\maketitle

\thispagestyle{firstpage}

\begin{abstract}

Deep Neural Networks (DNNs) and their accelerators are being deployed ever more frequently in safety-critical applications leading to increasing reliability concerns. A traditional and accurate method for assessing DNNs' reliability has been resorting to fault injection, which, however, suffers from prohibitive time complexity. While analytical and hybrid fault injection-/analytical-based methods have been proposed, they are either inaccurate or specific to particular accelerator architectures. 

In this work, we propose a novel accurate, fine-grain, metric-oriented, and accelerator-agnostic method called DeepVigor that provides vulnerability value ranges for DNN neurons' outputs. An outcome of DeepVigor is an analytical model representing vulnerable and non-vulnerable ranges for each neuron that can be exploited to develop different techniques for improving DNNs' reliability. Moreover, DeepVigor provides reliability assessment metrics based on vulnerability factors for bits, neurons, and layers using the vulnerability ranges.

The proposed method is not only faster than fault injection but also provides extensive and accurate information about the reliability of DNNs, independent from the accelerator. The experimental evaluations in the paper indicate that the proposed vulnerability ranges are 99.9\% to 100\% accurate even when evaluated on previously unseen test data. Also, it is shown that the obtained vulnerability factors represent the criticality of bits, neurons, and layers proficiently. DeepVigor is implemented in the PyTorch framework and validated on complex DNN benchmarks.

\end{abstract}

\blfootnote{The work is supported in part by the European Union through European Social Fund in the frames of the ``Information and Communication Technologies (ICT) programme'' (``ITA-IoIT'' topic), by the Estonian Research Council grant PUT PRG1467 ``CRASHLES'' and by Estonian-French PARROT project ``EnTrustED''.}

%\keywords{Deep Neural Networks, Reliability Assessment, Vulnerability, Resilience Analysis}
%\begin{IEEEkeywords}
%Deep Neural Networks, Reliability %Assessment, Vulnerability, Resilience %Analysis
%\end{IEEEkeywords}

\section{Introduction}
Deep Neural Networks (DNNs) have recently emerged to be exploited in a wide range of applications. DNN accelerators have also penetrated into safety-critical applications e.g., autonomous vehicles \cite{bosio2021emerging,forsberg2020challenges}. Therefore, several concerns are raised regarding developing and utilizing DNN accelerators in the realm of safety-critical applications, one of them being the reliability. 

\begin{figure}[t]
    \includegraphics[width=0.35\textwidth]{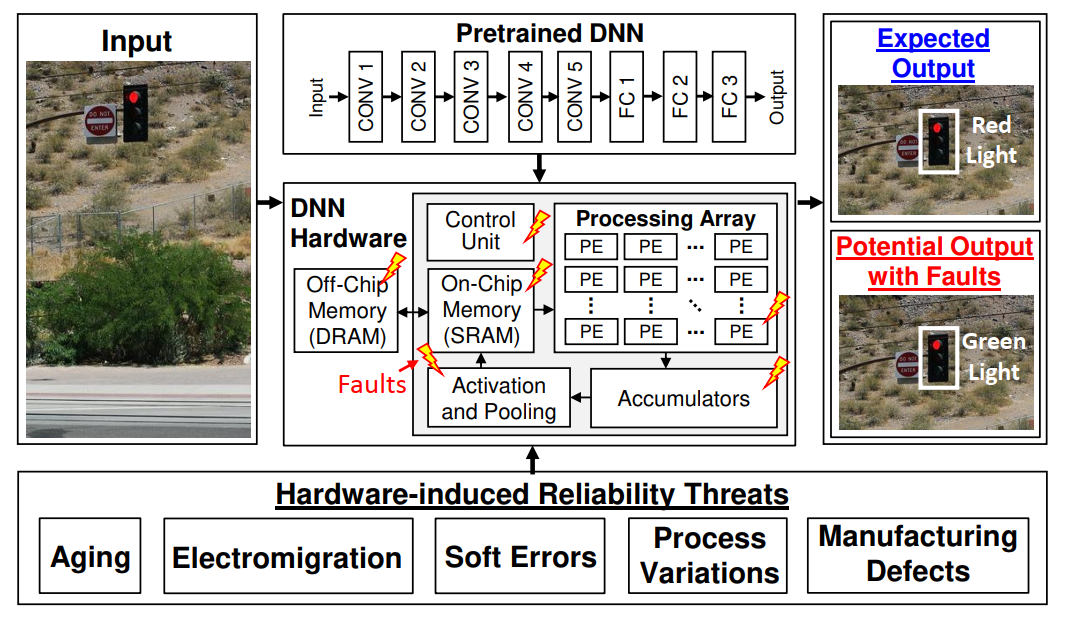}
    \centering
    \caption {Hardware reliability threats in DNN accelerators and their impact on the output \cite{bosio2021emerging}.}
    \label{fig:rel-thr}
\end{figure}

Reliability of DNNs concerns their accelerators' ability to perform correctly in the presence of faults \cite{ibrahim2020soft} originating from either the environment (e.g., soft errors, electromagnetic effects, temperature variations) or inside of the chip (e.g., manufacturing defects, process variations, aging effects) \cite{bosio2021emerging,shafique2020robust}. As shown in Fig. \ref{fig:rel-thr}, faults may occur in different locations of accelerators either in memory or logic components and they influence the parameters (e.g., weights and bias) and intermediate results (layers' activations) of neural networks that can decrease their accuracy drastically \cite{li2020soft,neggaz2019cnns}. By technology miniaturization, the effect of Single Event Transient (SET) and Single Event Upset (SEU) faults in devices is increasing thereby jeopardizing the reliability of modern digital systems \cite{azizimazreah2018tolerating}.

Recently, several works have been published on the assessment and improvement of the reliability of a variety of DNNs as well as on  different levels of system hierarchy \cite{ibrahim2020soft,shafique2020robust,mittal2020survey}. Reliability assessment is the process of modeling %or presenting 
the target DNN accelerator and measuring its reliability with respect to the corresponding quantitative evaluation metrics. Reliability assessment is the underlying procedure for improving reliability since it presents how the system could be influenced by threats as well as which locations of the system are more vulnerable to them. Therefore, it is the very first and principal phase of a reliable design process. 

Throughout the literature, reliability assessment methods for DNNs are mainly categorized into two major classes: fault injection (FI) and resilience analysis. The majority of the works assess the reliability of DNNs relying on FI, which provides realistic results on the impact of different fault models on the system's execution and is performed directly on the target platform (accelerator's software \cite{chen2021low} or RTL model \cite{xu2020hybrid}, FPGA \cite{xu2021reliability}, GPU \cite{basso2020impact}). FI outputs different evaluations for DNNs' reliability by accuracy loss, vulnerability factors, or fault classification \cite{xu2021reliability,dos2018analyzing,khoshavi2020shieldenn}. Moreover, fine-grain evaluations for finding critical bits can be performed by exhaustive FI or an optimized method in \cite{chen2019binfi}.

Nevertheless, FI methods are prohibitively time-consuming and carry a high complexity due to the need to inject an enormous amount of faults into a huge number of DNN parameters as well as time instances to reach an acceptable confidence level \cite{deepaxe,appraiser}. The more fine-grain evaluation is required the more sophisticated experiments should be performed. In addition, most faults in a FI experiment on DNNs are masked \cite{bosio2019reliability} and are thus unnecessarily examined. Furthermore, the outcome of such assessment is application/platform specific which can not be generalized for other platforms \cite{ruospo2021pros}.

Resilience analysis methods cope with the drawbacks of FI. They analyze the function of DNNs mathematically and have the potential to evaluate their reliability with arbitrary metrics. Therefore, resilience analysis methods can provide a deeper insight into the reliability evaluations of DNNs with lower complexity. Moreover, they can be conducted in different fault-tolerant designs on various platforms \cite{mahmoud2020hardnn}. 

Layer-wise Relevance Propagation (LRP) algorithm is leveraged in \cite{schorn2018accurate,schorn2019efficient,ruospo2021reliability,abdullah2020salvagednn} to obtain the contribution of neurons to the output to express their criticality and apply protections to improve the reliability of DNN accelerators. The sensitivity of DNN's filters is obtained by Taylor expansion with given error rates in \cite{choi2019sensitivity} for designing an error-resilient and energy-efficient accelerator. 

The conducted resilience analyses in these works are not able to provide reliability measurement metrics and detailed vulnerability evaluations. Moreover, they combine the criticality scores of neurons over individual outputs of the DNNs, thus resulting in missing important information about the resilience of DNNs as a whole. Mahmoud et al. \cite{mahmoud2020hardnn} proposed different heuristics for vulnerability estimation of feature maps without FI. These estimations which are more coarse grain than the LRP-based methods, lead to hardening the accelerators, however, the accuracy of the vulnerability estimation methods is remarkably lower than that of fault-injection methods.

The aforementioned papers on resilience analysis methods have focused mainly on finding the most critical neurons/weights in a DNN to protect them against faults in a fault-tolerant design. In addition, they do not explain sufficiently how a fault propagates through the network and influence its outputs. Fidelity framework \cite{he2020fidelity} is proposed to take advantage of both FI and analyzing DNN accelerators to provide reliability metrics. However, it requires detailed information of the accelerator architecture/implementation. To the best of our knowledge, there is no accelerator-agnostic resilience analysis method for DNNs that can compete with FI in terms of reliability evaluation to be less time-consuming, and accurate with fine-grain metrics enabling different reliability improvement techniques. 

In this research work, we introduce the concept of neurons' vulnerability ranges expressing whether or not a fault at the output of neurons would misclassify the network. Thus, it enables a comprehensive reliability study with a novel resilience analysis method called DeepVigor where the vulnerability factors of layers, neurons, and bits in a DNN are obtained. The contributions in this work are:

\begin{itemize}
    \item Proposing DeepVigor, a novel accurate, metric-oriented, and accelerator-agnostic resilience analysis method for DNNs reliability assessment faster than fault injection;
    \item Introducing and acquiring vulnerability ranges for all neurons in DNNs, assisted by a fault propagation analysis, providing accurate categorization of critical/non-critical faults;
    \item Providing fine-grain vulnerability factors as reliability evaluation metrics for layers, neurons, and bits in DNNs, compared with and validated by fault injection.
\end{itemize}

The remainder of the paper is organized as follows: the resilience analysis method is presented in Section \ref{method}, and the experimental setup and results are provided in Section \ref{results}. The applicability of the method is discussed in Section \ref{discussion}, and the work is concluded in Section \ref{conclusion}.

\section{DNN Reliability Assessment with DeepVigor} \label{method}

\subsection{Fault Model}

In this work, the fault propagation analysis is performed at the outputs of DNN neurons. However, they will cover a vast majority of internal faults of the neurons occurring inside the MAC units and also a large portion of faults in the weights and neurons' input activations. It is assumed that only one neuron has an erroneous output per execution due to faults which is a common assumption in the literature \cite{chen2019binfi}.

For validation by FI, the single-bit fault model has been applied. While the multiple-bit fault model is more accurate, it requires a prohibitively large number of fault combinations to be considered ($3^n - 1$ combinations, where $n$ is the number of bits). Fortunately, it has been shown that high fault coverage obtained using the single-bit model results in a high fault coverage of multiple-bit faults \cite{bushnell2004essentials}. Therefore, a vast majority of practical FI and test methods are based on the single-bit fault assumption. Single bitflip faults are injected randomly at neurons' outputs and once per execution. 

\subsection{Fault Propagation Analysis}

\begin{figure}[h]
    \includegraphics[width=0.35\textwidth]{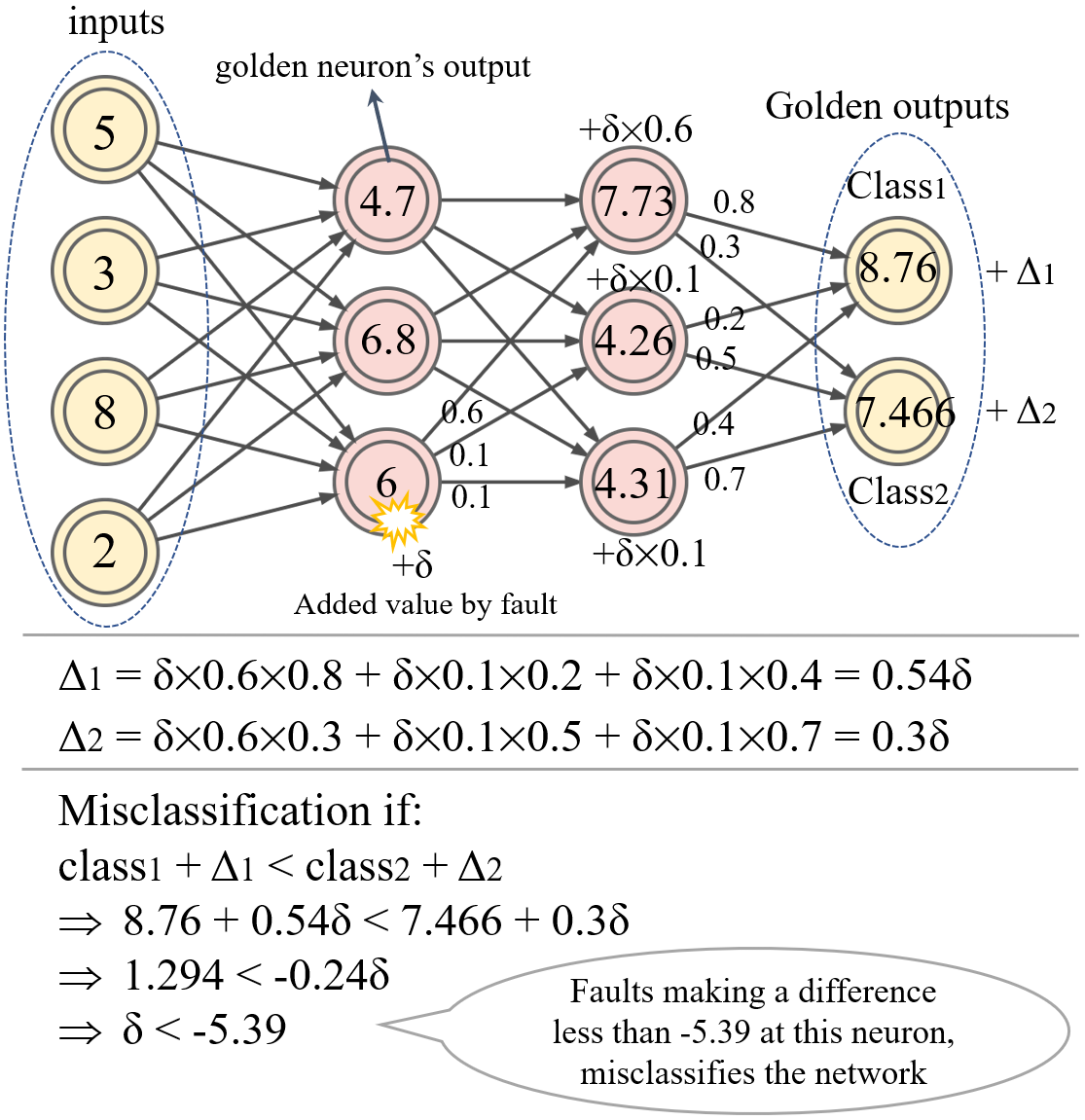}
    \centering
    \caption {An example of fault propagation analysis model and finding the vulnerability value ranges for a neuron with a given input.}
    \label{fig:idea-abstract}
\end{figure}

Fig.~\ref{fig:idea-abstract} depicts an overview of the rationale behind the DeepVigor method. A tiny neural network with few layers and neurons with given inputs, golden (fault-free) activation values (inside of neurons), and weights (on the arrows) is shown. The golden classification output is \textit{class1}. A fault changes the neuron's output by $\delta$ which is the difference between the golden and faulty activation values. This $\delta$ that can have either a negative or a positive value will be propagated to the output layer and may change the classification result. The fault propagation will make a difference on each output class as $\Delta_1$ and $\Delta_2$. Misclassification happens when the value of the output neuron \textit{class2} gets higher than that of neuron \textit{class1}. 

Thus, the propagation of the fault can be traced from the neuron to the output and a problem for misclassification can be expressed as shown in Fig.~\ref{fig:idea-abstract}. By solving the problem of misclassification condition in the output, the value for $\delta$ is obtained as a vulnerability threshold that expresses how much a fault should influence the neuron to misclassify the network. Therefore, a vulnerability value range for the neuron is acquired. In this example, the range $(-\infty, -5.39)$ is a vulnerable range and $[-5.39, +\infty)$ is non-vulnerable range. This idea is generalized for a DNN including multiple output classes and other corresponding functions in this paper.

\subsection{The DeepVigor Method} \label{method-steps}

The steps of the proposed DNNs' resilience analysis method (DeepVigor) and its validation are illustrated in Fig.~\ref{fig:Analysis-steps}. As shown, an analysis is performed on a set of data (i.e., set1, training set) and outputs the vulnerability value ranges as well as the vulnerability factors. Furthermore, FI is performed on the same and different data (i.e., set2, test set) to validate the outcomes of the analysis. The steps of DeepVigor are as follows: 

\begin{figure}[h]
    \includegraphics[width=0.4\textwidth]{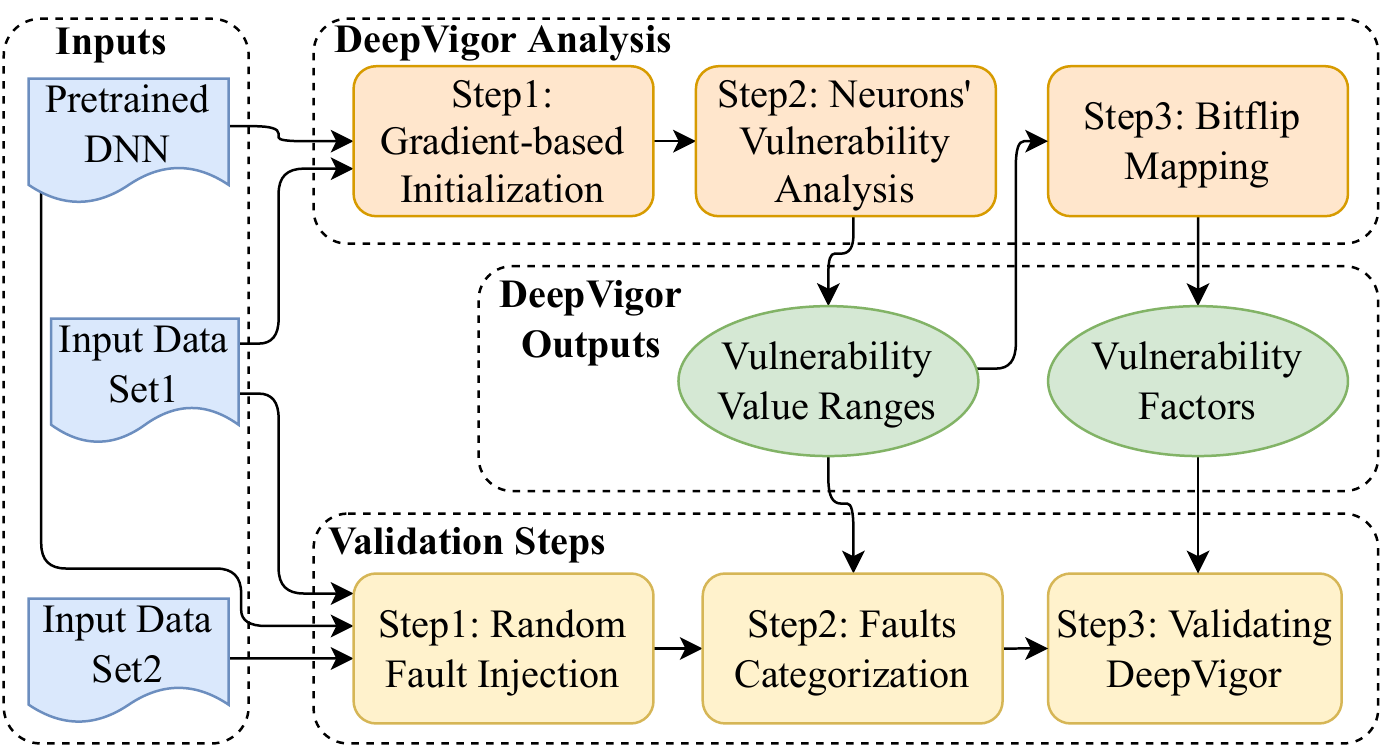}
    \centering
    \caption {Steps of the DeepVigor method for DNNs' reliability assessment and its validation.}
    \label{fig:Analysis-steps}
\end{figure}

\textbf{Step1 - Gradient-based Initialization:}
In the first step, a neuron is examined whether or not to be processed for the vulnerability analysis. For this purpose, assuming a neural network consisting of $L$ layers with $N$ output classes in $C = \{c_1, c_2, ..., c_N\}$. Neuron $k$ at layer $l$ is selected to be examined. The neuron's output is corrupted by adding a sample positive or negative value as $\epsilon_k^l$ to its output and the feed-forward of the network is executed over a batch of input data. A loss function $\mathcal{L}$ is defined in Equation \eqref{eq:loss-func} as: 
\begin{equation}
    \mathcal{L} = sigmoid(\sum_{j=0}^{N}(\mathcal{E}_{c_t} - \mathcal{E}_{c_i}))
    \label{eq:loss-func}
\end{equation}
where $c_t$ is the golden top class and $\mathcal{E}_{c_t}$ and $\mathcal{E}_{c_i}$ are the erroneous output values corresponding to the respective classes. The loss function computes the summation of differences between the value of the golden top class and the other outputs in the corrupted network and applies a $sigmoid$ function. The \textit{golden top class} is what the fault-free DNN outputs as its classification whether or not it is correctly classified.

$\mathcal{L}$ represents the impact of the neuron's erroneous output on the golden top class of the network. When the gradient of $\mathcal{L}$ w.r.t. the corrupted neuron's output for one input is zero, it means that any error at this neuron's output does not change the output classification. Considering a batch of inputs, if the gradients are zero for a portion of inputs larger than a threshold, the neuron is disregarded for the vulnerability analysis. In case most of the gradients are not zero, a range for searching the vulnerability value is initialized.

Considering $\epsilon_k^l$ is a positive value for one input, in case the gradient is positive, there is a minimum value $0 < \delta_k^l < \epsilon_k^l$ for the neuron that if error $\delta_k^l$ is added to its output (by a fault at its inputs or the output value itself) the network's golden classification would change. But if the gradient is negative, then $\delta_k^l$ should be searched through the values larger than $\epsilon_k^l$. A similar scenario is valid for negative values of $\epsilon_k^l$.

\textbf{Step2 - Neurons' Vulnerability Analysis:} 
In this step, the vulnerability ranges of neurons under analysis are obtained. 
Let $R_{NV}(l,k,x) = [r_{lower},r_{upper}]$ be a \emph{Range of Non-vulnerable Values} for a $k$-th neuron at layer $l$ with input data $x$. The bounds of range $R$ for $x$ are calculated as follows:
    
\begin{equation}
    \begin{split}
        \begin{cases}
            r_{upper} = min(\delta_k^l), \delta_k^l>0 , \mathcal{E}_{c_t} <  \mathcal{E}_{c_i}, i \neq t \\
            r_{lower} = max(\delta_k^l), \delta_k^l<0 , \mathcal{E}_{c_t} <  \mathcal{E}_{c_i}, i \neq t
        \end{cases}
    \end{split}
    \label{eq:analysis_eqs}
\end{equation}
where $c_t$ and $c_i$ are the golden top class and any other output class, respectively, and $\mathcal{E}_{c_t}$ and $\mathcal{E}_{c_i}$ are the erroneous output values corresponding to the respective classes.

Equation \eqref{eq:analysis_eqs} finds the maximum negative and minimum positive values induced at the corresponding neuron that do not lead to misclassifying the input data from the golden classification. Further, a \emph{Range of Vulnerable Values} $R_{VV}(l,k,x)$ for a $k$-th neuron at layer $l$ with input data $x$ is equal to $R_{VV} = (-\infty,r_{lower}) \cup (r_{upper},\infty)$.

Note, the equation is applied for a single input data. In the case of a data set $X$ containing $T$ input data $x_j$ the $R_{NV}$ and $R_{VV}$ will get refined and will be equal to intersections of their respective ranges over all inputs $x_j$ as follows:
\begin{equation}
    \begin{split}
        \begin{cases}
            R_{NV}(l,k) = \bigcap\limits^T_{j=1}R_{NV}(l,k,x_j) \\
            R_{VV}(l,k) = \bigcap\limits^T_{j=1}R_{VV}(l,k,x_j)
        \end{cases}
    \end{split}
    \label{eq:intersec}
\end{equation}

\begin{figure}[h]
    \includegraphics[width=0.4\textwidth]{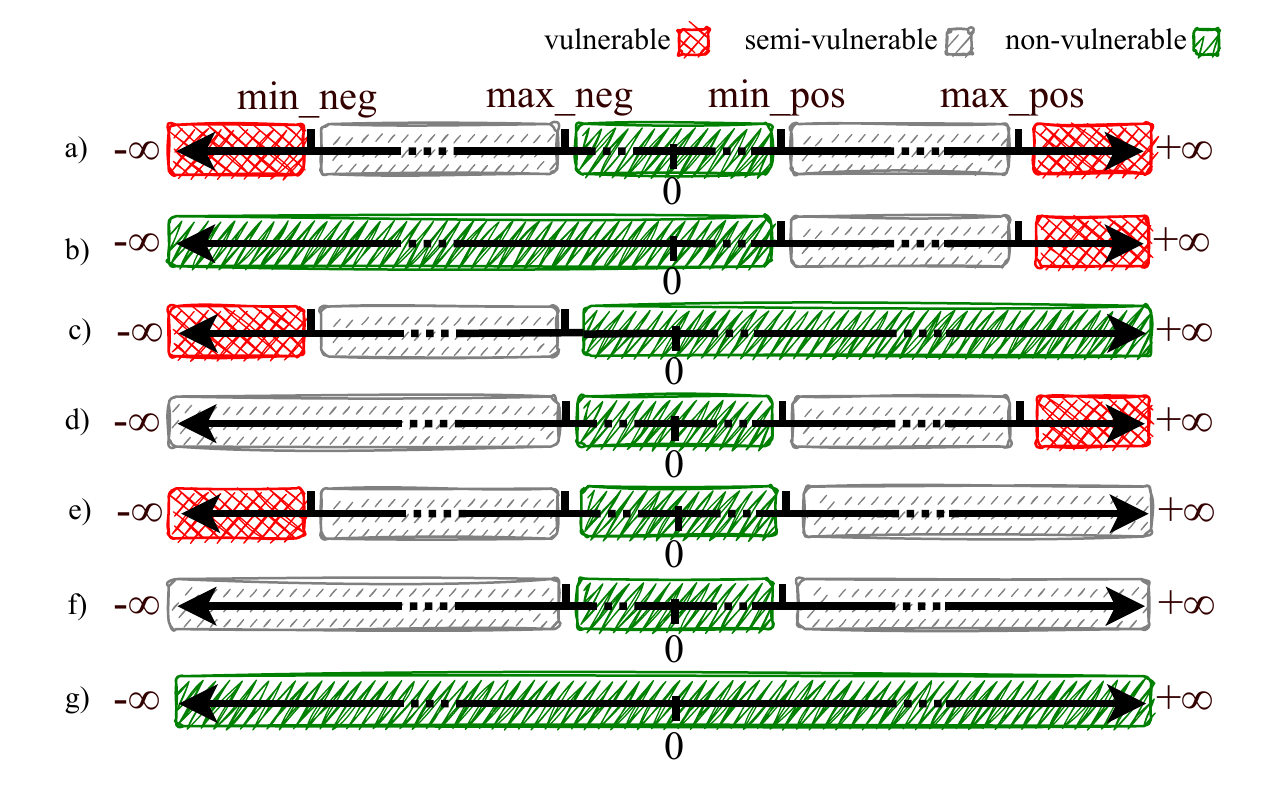}
    \centering
    \caption {Different possible cases of vulnerability ranges for each class in a neuron.}
%    \caption {Different possible cases of vulnerability ranges for each class in a neuron: a) all vulnerability ranges present, b) $[-\infty, min\_pos]$ is non-vulnerable, $[min\_pos, max\_pos]$ is semi-vulnerable, $[max\_pos, +\infty]$ is vulnerable, c) mirror case of b, d) $[-\infty, max\_neg]$ and $[min\_pos, max\_pos]$ are semi-vulnerable, $[max\_neg, min\_pos]$ is non-vulnerable, $[max\_pos, +\infty]$ is vulnerable, e mirror case of d, f) $[-\infty, max\_neg]$ and $[min\_pos, +\infty]$ are semi-vulnerable, $[max\_neg, min\_pos]$ is non-vulnerable, g) all numbers are non-vulnerable.}
    \label{fig:vul-ranges}
\end{figure}

The outcome of solving the equations for each neuron and merging the results over all inputs will be the vulnerability value ranges for each class separately, each range specifies the impact of a fault on changing the neuron value whether it influences the network classification result or not. Fig. \ref{fig:vul-ranges} depicts different cases for vulnerability ranges over all numbers. Three vulnerability ranges are identified as follows:

\begin{itemize}
    \item \textbf{Non-vulnerable range:} If a fault lay an effect on the neuron output in this range, no misclassification happens (hachured-green sections in Fig. \ref{fig:vul-ranges});
    \item \textbf{Vulnerable range:} If a fault makes a difference at the output of the neuron in this range, the output will be misclassified (cross hachured-red sections in Fig. \ref{fig:vul-ranges});
    \item \textbf{Semi-vulnerable range:} If a fault causes the neuron value to move as an amount in this range, this fault \textit{may} cause a misclassification (dashed-grey sections in Fig. \ref{fig:vul-ranges}). Cases \textit{d}-\textit{f} in Fig. \ref{fig:vul-ranges} happen when the portion of zero gradients in \textit{step1} is less than the \textit{threshold} and more than $\textit{1}-threshold$.
    
\end{itemize}

\textbf{Step3 - Bitflip Mapping:}
In this step, DeepVigor maps the neurons' bitflipped values over input data on the vulnerability value ranges to indicate fine-grain vulnerability factors as metrics for the DNNs' reliability. For this purpose, the inputs used in \textit{step2} and obtained vulnerability value ranges are fed to the network and in each bit of each neuron, bitflips are performed. In each bitflip, the difference in the new value of the target neuron is calculated and compared with the corresponding vulnerability range. 

Based on the range of what the bitflip maps, the bit is considered vulnerable or non-vulnerable, respectively. By this analysis, the number of vulnerable bits of the neurons is obtained over the inputs. Hence, vulnerability factors of each layer (LVF), neuron (NVF), or bit (BVF) of the DNN can be defined as equations \eqref{eq:critical-layer}, \eqref{eq:critical-neuron}, and \eqref{eq:critical-bit}, respectively. Vulnerability factors express the probability of misclassifying the network in case of the occurrence of a bitflip at the target element.

\begin{equation} 
    \begin{split}
        LVF  = \qquad \qquad \qquad \qquad \qquad \qquad \qquad \qquad \qquad \qquad
        \\ \frac{\#vulnerable\;bits\;in\;layer}{\#inputs\; \times \#layer's\; neurons \times word\;length} \times 100
    \end{split}
    \label{eq:critical-layer}
\end{equation}

\begin{equation}
    \begin{split}
        NVF = \frac{\#vulnerable\;bits\;in\;neuron}{\#inputs\; \times word\;length} \times 100
        \end{split}
        \label{eq:critical-neuron}
\end{equation}

\begin{equation}
    \begin{split}
        BVF = \frac{\#vulnerable\;times\;for\;bit}{\#inputs\; \;} \times 100
    \end{split}
    \label{eq:critical-bit}
\end{equation} 

\subsection{Validating DeepVigor By Fault Injection}
As illustrated in Fig.~\ref{fig:Analysis-steps}, DeepVigor results are validated by means of FI over the input data and categorizing faults based on the vulnerability value ranges. The steps of the validation process of DeepVigor are as follows:

\textbf{Step1 - Random Fault Injection:} 
According to the adopted fault model, when one input is fed to the network, a random single bitflip is injected into a random neuron in a layer. This process is repeated several times for one input depending on the number of neurons and word length of data to reach a 95\% confidence level and 1\% error margin based on \cite{leveugle2009statistical}. The required number of faults is obtained by Equation \eqref{eq:statistical-fi} where $N = word \; length \times \#layer's\; neurons$ that represents the total number of bits in the output of a layer.

\begin{equation}
    \begin{split}
        \#layer's \; random \; faults = \frac{N}{1 + (0.01^2 \times \frac{N - 1}{1.96^2 \times 0.5^2})}
    \end{split}
    \label{eq:statistical-fi}
\end{equation} 

\textbf{Step2 - Fault Categorization:} 
Once a fault is injected, a difference is produced in the output of the neuron in comparison with the golden model. In this step, the produced difference by a fault at the neuron's output is compared with the obtained vulnerability ranges, and faults are categorized as:

\begin{itemize}
    \item \textbf{Non-critical fault:} The produced difference is in the non-vulnerable range.
    \item \textbf{Critical fault:} The produced difference is in the vulnerable range.
\end{itemize}

\textbf{Step3 - Validating DeepVigor:}
To validate DeepVigor by FI, injected faults are propagated to the output and the network classification output is examined. The accuracy of the method is defined based on the two metrics as follows:

\begin{itemize}
    \item \textbf{True non-critical faults:} Percentage of faults that are categorized as non-critical and do not change the classification at the output;
    \item \textbf{True critical faults:} Percentage of faults that are categorized as critical and change the classification at the output.
\end{itemize}

Another metric for validating the outputs of DeepVigor is the correlation between LVF and DNN's accuracy loss. This correlation shows that the obtained vulnerability factors from DeepVigor represent the criticality of the components properly.
Since other vulnerability factors (NVF and BVF) are calculated using the same vulnerability ranges, by validating LVF, they will be also liable metrics for the resilience analysis, consequently.

\section{Experimental Results} \label{results}

\subsection{Experimental Setup}

All DNNs, steps of DeepVigor, and its validation are implemented in PyTorch and run on NVIDIA 3090 GPU. To explore different DNN structures, six representative DNNs trained on three datasets are examined for the experimental results. We have experimented with two 5-layer MLPs (one with Sigmoid and one with ReLU) trained on MNIST, two LeNet-5 with 3 convolutional (CONV) layers, 2 max-pooling (POOL) layers, and 2 fully-connected (FC) layers trained on MNIST and CIFAR-10, AlexNet with 5 CONV, 3 POOLs, 2 batch normalization (BN) and 3 FCs trained on CIFAR-10, and VGG-16 with 13 CONV, 13 BNs, 5 POOLs and 2 FCs trained on CIFAR-100. The respective networks' accuracy on the corresponding test sets are 94.64\%, 90.55\%, 90.4\%, 66.15\%, 72.73\%, and 69.41\%.

Data representation in this work is 32-bit floating point IEEE-754 and the $word \; length$ in equations \eqref{eq:critical-layer}-\eqref{eq:statistical-fi} is 32 bits. For validation, a layer-wise statistical random FI is performed that satisfies a 95\% confidence level and 1\% error margin.

In the first step of DeepVigor $\epsilon_k^l$ is considered $-/+10000$ for range initialization and the whole search range is $[-5 \times 10^{5}, 5 \times 10^{5}]$. Finding $\delta_k^l$ in all networks by a logarithmic search is performed for negative and positive numbers separately, considering a 0.05 difference from the main value. Also, based on empirical explorations the threshold of neurons' zero-gradients for inputs is considered 98\% for all experiments. Corresponding experiments are performed on the whole sets of training (as the input data set1) and test (as the input data set2) data.

\subsection{Results and Validation}

We analyze all neurons of the representative DNNs with training sets as the input data set1 by DeepVigor and obtain the vulnerability ranges. In the fault categorization step, faults are categorized into critical and non-critical classes with an accuracy close to 100\%. Throughout the results from FI experiments, DeepVigor identified 66.63\% to 99.42\% of faults as non-critical over different layers of analyzed networks. 

For validation, Table~\ref{tab:val-results-train} presents the range of obtained accuracy values of the method through all layers of DNNs in terms of true non-critical and critical faults. It is observed that the accuracy of the method for categorizing non-critical faults is 99.950\% to 100\% and for critical faults ranging from 99.955\% to 100\% for the same data set. 

\begin{table}[t]
\small
\centering
\caption{Accuracy of DeepVigor by fault injection on the same input data as the analysis.}
\resizebox{0.45\textwidth}{!}{%
\begin{tabular}{|c|c|c|}
\hline
DNN             & True non-critical faults & True critical faults \\ \hline
MLP-sigmoid-mnist   & 99.985\%$\sim$100\%      & 100\%            \\ \hline
MLP-relu-mnist  & 99.991\%$\sim$100\%      & 100\%                \\ \hline
LeNet-mnist     & 99.992\%$\sim$100\%        & 100\%                \\ \hline
LeNet-cifar10   & 99.956\%$\sim$100\%       &  100\%  \\ \hline
AlexNet-cifar10 & 99.973\%$\sim$100\%      & 99.955\%$\sim$100\%  \\ \hline
VGG16-cifar100  & 99.950\%$\sim$100\%       & 99.972\%$\sim$100\%  \\ \hline
\end{tabular}%
}
\label{tab:val-results-train}
\end{table}

The minor error seen in the results is due to: 1) Considered error in finding vulnerability values, 2) FI results in "NaN" values in 32-bit floating point IEEE-754 while the computations are being done on a GPU. We have categorized them as critical faults, 3) the effect of few inputs with non-zero gradients in \textit{step1} as described in \ref{method-steps}.

We have also experimented with FI on the test sets (input data set2) to see the validity of the analysis on different sets reported in Table \ref{tab:val-results-test}. As it can be seen, similar high accuracy values to input data set1 are obtained.

\begin{table}[h]
\small
\centering
\caption{Accuracy of DeepVigor by fault injection on a different input data from the analysis.}
\resizebox{0.45\textwidth}{!}{%
\begin{tabular}{|c|c|c|}
\hline
DNN             & True non-critical faults & True critical faults   \\ \hline
MLP-sig-mnist   & 99.985\%$\sim$99.996\%   & 99.911\%$\sim$100\%    \\ \hline
MLP-relu-mnist  & 99.976\%$\sim$100\%      & 100\%                  \\ \hline
LeNet-mnist     & 99.992\%$\sim$100\%        & 100\%                  \\ \hline
LeNet-cifar10   & 99.952\%$\sim$100\%      & 99.970\%$\sim$100\%     \\ \hline
AlexNet-cifar10 & 99.951\%$\sim$99.997\%   & 99.948\%$\sim$99.998\% \\ \hline
VGG16-cifar100  & 99.950\%$\sim$99.983\%    & 99.972\%$\sim$99.998\% \\ \hline
\end{tabular}%
}
\label{tab:val-results-test}
\end{table}

\begin{figure}[h]
\begin{tikzpicture}
\tikzstyle{every node}=[font=\small]
 \pgfplotsset{
  y ticks with fixed point/.style={
      yticklabel={
        \pgfkeys{/pgf/fpu=true}
        \pgfmathparse{exp(\tick)}%
        \pgfmathprintnumber[fixed relative, precision=3]{\pgfmathresult}
        \pgfkeys{/pgf/fpu=false}
      }
  }
 }
  \begin{axis}[
        width=\columnwidth,
        height=0.5\columnwidth,
        scaled y ticks = false,
        xtick={1,2,3,4,5,6,7},
        ytick={0,0.5,1,1.5,2,2.5,3,3.5,4},
        xticklabels = {conv1, conv2, conv3, conv4, conv5, fc1, fc2},
        yticklabels = {\strut $0.0$,\strut $0.5$,\strut $1.0$,\strut $1.5$,\strut $2.0$,\strut $2.5$,\strut $3.0$, \strut $3.5$,\strut $4.0$},
        ymin=0, ymax=4,
        xmin=0.5, xmax=7.5,
         grid=major,
         grid style={dashed,gray}, 
        ylabel near ticks,
        xlabel near ticks,
        xlabel= Layers of AlexNet, 
        ylabel= LVF (\%),
        legend columns = 3,
        legend style={at={(1,1)},anchor=north},
        legend style={draw=black, at={(0.5,1)}, text opacity = 1,row sep=0pt, nodes={scale=0.6}},
         x tick label style={rotate=0,anchor=north},
        ]
        
        \addplot [blue, thick ,mark=o , mark size=2pt] table [x=layers, y=acc loss same, col sep=comma] {images-ETS/alexnet-lvf.csv};
         \addplot [green, thick,mark=triangle*,mark size=2pt] table [x=layers, y=acc loss diffr, col sep=comma] {images-ETS/alexnet-lvf.csv};
         \addplot [red, thick ,mark=square , mark size=2pt] table [x=layers, y=LVF, col sep=comma] {images-ETS/alexnet-lvf.csv};
    
        \legend{LVF, Accuracy Loss (same sets), Accuracy Loss (different sets)}
      \end{axis}
    \end{tikzpicture}
\centering 
     \caption {Correlation between LVF and accuracy loss.}
     \label{fig:lvf-acc}
 \end{figure}
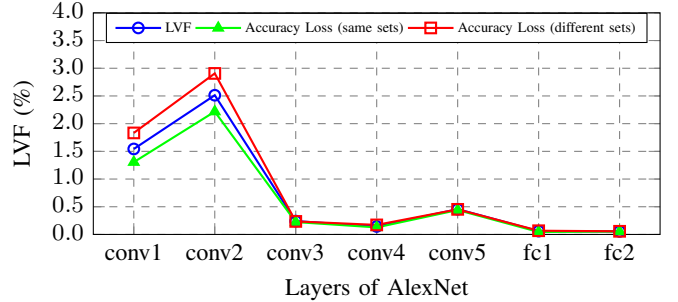

To validate the vulnerability factors, Fig.~\ref{fig:lvf-acc} illustrates the correlation between LVF and accuracy loss for a layer-wise FI on AlexNet. As demonstrated, there is a close relationship between the LVF obtained from DeepVigor and accuracy loss in FI, either the input sets are similar or different.
This correlation is observed similarly in the results for all experimented DNNs. Therefore, LVF represents the vulnerability of layers competently. 

% \begin{figure}[h]
%     \includegraphics[width=0.45\textwidth]{images/lvf-accloss.png}
%     \centering
%     \caption {Correlation between LVF (obtained from DeepVigor on input data set1) and accuracy loss (obtained from layer-wise FI into input data set1 and set2).}
%     \label{fig:lvf-acc}
% \end{figure}

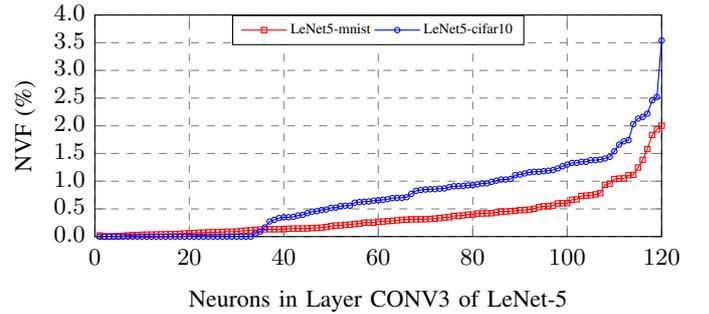
\begin{figure}[h]
\begin{tikzpicture}

\tikzstyle{every node}=[font=\small]
\pgfplotsset{
  x ticks with fixed point/.style={
      xticklabel={
        \pgfkeys{/pgf/fpu=true}
        \pgfmathparse{exp(\tick)}%
        \pgfmathprintnumber[fixed relative, precision=3]{\pgfmathresult}
        \pgfkeys{/pgf/fpu=false}
      }
  },
  y ticks with fixed point/.style={
      yticklabel={
        \pgfkeys{/pgf/fpu=true}
        \pgfmathparse{exp(\tick)}%
        \pgfmathprintnumber[fixed relative, precision=3]{\pgfmathresult}
        \pgfkeys{/pgf/fpu=false}
      }
  }
}
  \begin{axis}[
        width=\columnwidth,
        height=0.5\columnwidth,
        scaled x ticks = false,
        scaled y ticks = false,
        xtick={0,20,40,60,80,100,120}, 
        ytick={0,0.5,1,1.5,2,2.5,3,3.5,4},
        xticklabels = {\strut  $0$,\strut $20$,\strut $40$,\strut $60$,\strut $80$,\strut $100$, \strut$120$},
        yticklabels = {\strut $0.0$,\strut $0.5$,\strut $1.0$,\strut $1.5$,\strut $2.0$,\strut $2.5$,\strut $3.0$, \strut $3.5$,\strut $4.0$},
        ymin=0, ymax=4,
        xmin =0, xmax=120,
        grid=major, 
        grid style={dashed,gray}, 
        ylabel near ticks,
        xlabel near ticks,
        ylabel= NVF (\%), 
        xlabel= Neurons in Layer CONV3 of LeNet-5,
        legend columns = 3,
        legend style={at={(1,1)},anchor=north},
        legend style={draw=black, at={(0.5,1)}, text opacity = 1,row sep=0pt, nodes={scale=0.6}},
         x tick label style={rotate=0,anchor=north},
        ]
        \addplot [red, mark=square , mark size=1pt] table [x=neurons, y=LeNet5-mnist, col sep=comma] {images-ETS/lenet-nvf.csv};
        \addplot [blue,mark=o , mark size=1pt] table [x=neurons, y=LeNet5-cifar10, col sep=comma] {images-ETS/lenet-nvf.csv};
    
        \legend{LeNet5-mnist, LeNet5-cifar10}
      \end{axis}
    \end{tikzpicture}
    \centering
    \caption {NVF of neurons in CONV3 for LeNet5-mnist and LeNet-cifar10.}
    \label{fig:ncs-mnist}
\end{figure}

DeepVigor also provides \textit{NVF} and \textit{BVF} metrics as vulnerability factors for neurons and bits, respectively. As a representative example, Fig.~\ref{fig:ncs-mnist} depicts \textit{NVF} for layer \textit{conv3} of LeNet5-mnist and LeNet5-cifar10 that the more vulnerable neurons can be identified. In this figure, the number of neurons is sorted in each DNN separately, in the ascending order of NVF.  Also, \textit{BVF} for all neurons in DNNs is obtained and the results show that the most significant bit of exponents is the most vulnerable bit in most cases. 

\vspace{-10pt}

\vspace{-20pt}

\subsection{Run-Time Analysis}
DeepVigor enables a fine-grain reliability evaluation for DNNs faster than exhaustive FI. In our experiments, \textit{step1} of DeepVigor have removed up to 48\% of neurons' vulnerability analysis to be processed in \textit{step2}. Moreover, the range initialization in \textit{step1} has accelerated the search for finding the vulnerability values for 50\% to 99\% of neurons in \textit{step2} among the DNNs. Based on our experiments, a complete vulnerability range (as in Fig. \ref{fig:vul-ranges}) for one neuron can be obtained by 9.1 times feed-forward execution per neuron on average. While an exhaustive FI experiment runs the feed-forward by the number of bits (32 in our case) per neuron. Therefore, DeepVigor requires 3.5 times fewer feed-forwards translating into a similar amount of speed-up in run-time.

The run-time of DeepVigor depends on:
\begin{itemize}
    \item Backpropagation execution by the number of neurons \textit{step1} (one for positive and one negative numbers per neuron);
    \item Feed-forward execution by the number of searches for finding a positive or negative $\delta_k^l$ per neuron, in which the best case is 0 searches (in case of zero gradients), the moderate case is 14 searches (in case of limited range initialization), and the worst case is 22 searches;
    \item Vulnerability analysis of the neurons in the last layer is performed by simplified mathematics similar to Fig. \ref{fig:idea-abstract} and requires no iterative feed-forward or searching process through a wide range of numbers;
    \item Bitflip mapping is merely performing a bitflip at each neuron and a comparison with the obtained vulnerability ranges.
\end{itemize}

\section{Discussion} \label{discussion}

DeepVigor method is validated in the previous section, and it is shown how it can evaluate the reliability of DNNs proficiently with shorter run-times than FI. Vulnerability ranges enable a fine-grain and accurate resilience evaluation for neural networks. They are not limited to representing the single bitflip fault model and the outcome of the analysis is valid for an erroneous output for the neurons covering several fault models. This method enables an accelerator-agnostic analysis for DNNs and results can be applied to different accelerators. 

The outputs of DeepVigor provide different possibilities for exploiting techniques of reliability improvement, including:
\begin{itemize}
    \item Selective bits/neurons/layers hardening in accelerators based on the obtained BVF/NVF/LVF metrics;
    \item Fault-aware mapping for neurons on the processing elements of accelerators as in \cite{schorn2018accurate,ruospo2021reliability};
    \item Applying range restriction for neurons' or layers' outputs for preventing faults propagation as in \cite{chen2021low,hoang2020ft,ghavami2022fitact}.
\end{itemize}

\section{Conclusions} \label{conclusion}

In this work, a novel resilience analysis method for DNNs reliability assessment named DeepVigor is proposed. The output of this method is the vulnerability value ranges for all neurons through the DNNs which result in vulnerability factors for all layers, neurons, and bits of the DNN, separately. The method is validated extensively by fault injection and its feasibility to categorize non-critical and critical faults on complex DNNs with 99.9\% to 100\% accuracy is demonstrated. Moreover, vulnerability factors obtained by the proposed analysis provide fine-grain criticality metrics for DNNs' components leading to different reliability improvement techniques. The DeepVigor method is very proficient in the evaluation and explanation of the reliability of DNNs with shorter run-times than fault injection.

%\section*{Acknowledgments}

\bibliographystyle{IEEEtran}
\bibliography{refs.bib}
%\printbibliography

\end{document}